\documentclass[letterpaper, 10 pt, conference]{ieeeconf}  

\IEEEoverridecommandlockouts                              

\newcommand{\mbm}[1]{\mbox{\boldmath $#1$}}

\usepackage{graphics} 
\usepackage{epsfig} 
\usepackage{amsmath} 
\usepackage{graphicx}
\usepackage{amssymb}
\usepackage{balance}
\usepackage{tikz}
\usetikzlibrary{shapes.geometric}
\usepackage{subfiles}
\usepackage{subcaption}
\usepackage{algorithm,algorithmic}
\usepackage[colorinlistoftodos]{todonotes}
\newcommand{\norm}[1]{\left\lVert#1\right\rVert}

\newcommand{\cmmnt}[1]{}

\title{\LARGE \bf
Safe-To-Explore State Spaces: Ensuring Safe Exploration in Policy Search with Hierarchical Task Optimization
 }

\author{Jens Lundell$^{*}$, Robert Krug$^{\ddag}$, Erik Schaffernicht$^{\dagger}$, Todor Stoyanov$^{\dagger}$, Ville Kyrki$^{*}$
	\thanks{$^{*}$Intelligent Robotics Group,
		Department of Electrical Engineering and Automation, Aalto University,
		Finland. \{\texttt{firstname.lastname\}{@}aalto.fi}}
  \thanks{$^{\dagger}$AASS Research Center, \"{O}rebro University, Sweden. \{\texttt{firstname.lastname\}{@}oru.se}}
       \thanks{$^{\ddag}$ Robotics, Learning and Perception lab, KTH Royal Institute of 
       Technology, Sweden. \texttt{rkrug{@}kth.se}}
       \thanks{This work was supported by Academy of Finland decision 314180 
       and by the 
       Swedish 
       Foundation for
       	Strategic Research (SSF).}%
	}

\begin{document}

\maketitle
\thispagestyle{empty}
\pagestyle{empty}

\begin{abstract}
Policy search reinforcement learning allows robots to acquire skills by themselves. However, the learning procedure is inherently unsafe as the robot has no a-priori way to predict the consequences of the exploratory actions it takes. Therefore, exploration can lead to collisions with the potential to harm the robot and/or the environment. In this work we address the safety aspect by constraining the exploration to happen in safe-to-explore state spaces. These are formed by decomposing target skills (e.g., grasping) into higher ranked sub-tasks (e.g., collision avoidance, joint limit avoidance) and lower ranked movement tasks (e.g., reaching). Sub-tasks are defined as concurrent controllers (policies) in different operational spaces together with associated Jacobians representing their joint-space mapping. Safety is ensured by only learning policies corresponding to lower ranked sub-tasks in the redundant null space of higher ranked ones. As a side benefit, learning in sub-manifolds of the state-space also facilitates sample efficiency. Reaching skills performed in simulation and grasping skills performed on a real robot validate the usefulness of the proposed approach. 
\end{abstract}

\section{Introduction}
\label{sec:introduction}

Real-time robot motion planning and control is a key ability for autonomous robots operating in
unstructured environments. In recent years, policy search has emerged as one of the most promising
approaches to enable simultaneous motion planning and execution. It is a branch of reinforcement
learning (RL) focusing on optimizing parameterized controllers. Policy search scales gracefully to
high dimensions \cite{schaal2006dynamic}, but typically requires handcrafted low-dimensional policy
parameterizations to learn in as few roll-outs as possible
\cite{peters2007reinforcement,ijspeert2002movement,kalakrishnan2011learning,feirstein2016reinforcement}. In addition, policy search
requires a robot to explore its environment, an inherently unsafe process as it can lead to dangerous situations, such
as collisions with obstacles, which might cause damage to the robot and/or the environment. Also, in many practical applications the number of samples needed to ensure successful skill learning is prohibitive.

Classical approaches overcome such limitations by enforcing conservative policy 
updates between iterations \cite{peters2008reinforcement,schulman2015trust} or 
by discouraging entering parts of the state space by imposing 
penalties~\cite{deisenroth2011learning}. These approaches, however, only limit 
the risk of collision but cannot completely remove it. Instead, we propose a 
method to implicitly form \textit{safe-to-explore state spaces} (STESS) that 
focuses exploratory actions in arbitrary operational spaces to a collision free 
subset of the original state space. This is accomplished by first decomposing a 
robotic skill (\textit{e.g.}, reach-to-grasp movements) into prioritized 
elemental tasks defined in potentially different operational spaces of choice 
together with a map (\textit{i.e.}, a
Jacobian) from operational space to joint space. Then, a whole-body real-time inverse kinematics
scheme~\cite{kanoun2011kinematic, escande2014hierarchical} is used to obtain the corresponding joint
velocities while ensuring that higher ranked tasks (\textit{e.g.}, obstacle/joint limit avoidance)
are prioritized over lower ranked movement tasks. Therefore we can limit exploration to obey a pre-defined  operational space hierarchy. The control laws corresponding to lower ranked movement tasks are learned via policy search.

\begin{figure}
	\centering
	\includegraphics[width=0.8\linewidth]{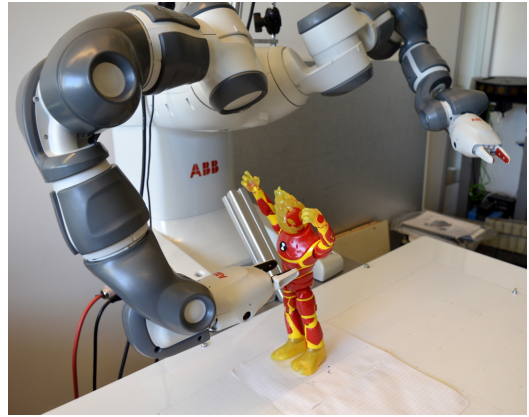}
	\caption{System setup consisting of an ABB YuMi robot and a toy to be grasped.}
	\label{fig:YuMi}
	  \vspace{-8mm}
\end{figure}

The main contributions of this work are:
\begin{itemize}
	\item a novel approach for ensuring safe exploration in policy search (Section \ref{sec:Methods}) consisting of a task-prioritized inverse kinematics framework (Section \ref{sec:grasping_system}) which facilitates safe learning of a time-invariant policy (Section \ref{sec:RL});	
	\item an experimental evaluation (Section \ref{sec:Exp_eval}) for a simulated reaching skill (Section \ref{sec:reaching_task}) and grasping
	skill on the platform shown in Fig.~\ref{fig:YuMi} (Section \ref{sec:grasping_task}) demonstrating that our method increases both safety, by removing the collision risk, and learning rate, by
	reducing the number of roll-outs before convergence. 
\end{itemize}

Per se, our method is not tied to a specific policy search algorithm. In this work, we use a time-invariant version of the 
model-free Policy Improvements with Weighted Exploration (PoWER) algorithm~\cite{kober2009policy}.


\section{Approach}
\label{sec:Methods}

Our approach consists of a hierarchical inverse kinematics (IK) whole body control framework and 
a policy search component. The aim of the framework is to produce joint-level commands by 
inverting a set of concurrent hierarchical controllers which operate in arbitrary operational 
spaces. In this work, we use policy search to learn a subset of lower ranked controllers 
responsible for movement generation. The policy is optimized with policy search where the goal 
is to learn policy parameters $\boldsymbol{\theta}$ that maximize the expected reward 
$E_{\pi_{\boldsymbol{\theta}}}\left\lbrack \sum_{t=1}^{T} r(\textbf{u}_t, 
\textbf{x}_t)\right\rbrack$ over trajectories 
$\tau=\left\lbrace\textbf{x}_1,\textbf{u}_1,\dots,\textbf{x}_T,\textbf{u}_{T-1}\right\rbrace$, 
where \textbf{x} are states, \textbf{u} are actions and expectations are taken with respect to 
the policy's trajectory distribution 
$p(\tau)=p(\mbm{x}_1)\prod_{t=1}^{T}p(\mbm{x}_{t+1}|\mbm{x}_t,\mbm{u}_t)p(\mbm{u_t}|\mbm{x}_t)$ 
where $p(\textbf{x}_{t+1}|\textbf{u}_t,\textbf{x}_t)$ represent system dynamics. In this work, 
actions are operational space velocities as the IK framework produces joint 
velocities, while states and 
rewards are application specific (Section \ref{sec:Exp_eval}).

\subsection{Task-prioritized inverse kinematics}
\label{sec:grasping_system}
The following task-prioritized real-time inverse kinematics scheme is described in depth in our
previous work~\cite{krug2016next}, which in turn builds upon the method developed
in~\cite{kanoun2011kinematic}. Here, we give a brief summary for completeness as it is an essential
component of the presented approach.

As in prior works~\cite{krug2016next,escande2014hierarchical}, we consider kinematic manipulator control with the goal of computing joint velocity commands $\dot{\mbm{q}}$ by inverting a set of hierarchical tasks defined in different operational spaces. The obtained joint velocities are then executed by a low-level tracking controller. We describe the joint configuration of a manipulator with the vector $\mbm{q}$. Following~\cite{kanoun2011kinematic}, we define task Jacobians via the derivatives of user-defined task functions $\mbm{e}(\mbm{q})$. To give an example, consider a static plane described by its unit normal $\mbm{n}$ and offset $d$. The task of moving an end-effector point $\mbm{p}(\mbm{q})$ onto this plane can be achieved by driving the residual of the function $\mbm{e}(\mbm{q})=\mbm{n}^T\mbm{p}(\mbm{q})-d$ to zero. Here, the corresponding operational space is defined by the normal $\mbm{n}$ and the Jacobian mapping from joint space to operational space is given by $\mbm{J}(\mbm{q})=\mbm{n}^T\frac{\partial \mbm{p}(\mbm{q})}{\partial \mbm{q}}$. A desired task evolution can be imposed via control laws (policies) $\dot{\mbm{e}}^*$ achieving, $e. \,g.$, exponential convergence by setting $\dot{\mbm{e}}^*=-\lambda \mbm{e}$ with $\lambda \in \mathbb{R}_+$. In the following, we drop dependencies on $\mbm{q}$ for notational convenience. For a single equality task, finding joint space controls corresponds to solving the following least-squares problem
\begin{align}
\arg \min_{\dot{\mbm{q}}} \; \frac{1}{2}\norm{\mbm{J}\dot{\mbm{q}}-\dot{\mbm{e}}^*}^2,
\label{eq:eq_task}
\end{align}
which can easily be done via a pseudoinverse of $\mbm{J}$. In order to allow 
for inequality tasks
and without loss of generality we henceforth use a general task formulation with upper bounds
\begin{align}
\mbm{J}\dot{\mbm{q}}\leq\dot{\mbm{e}}^*.
\label{eq:ineq_task}
\end{align}
As shown in~\cite{kanoun2011kinematic}, this allows for lower bounds
\mbox{$\mbm{J}\dot{\mbm{q}}\geq\dot{\mbm{e}}^*$}, double bounds \mbox{$
\dot{\underline{e}}^*\leq\mbm{J}\dot{\mbm{q}}\leq\dot{\overline{\mbm{e}}}^*$}, and equalities
\mbox{$\mbm{J}\dot{\mbm{q}}=\dot{\mbm{e}}^*$}, by  reformulating them, respectively, as
\mbox{$-\mbm{J}\dot{\mbm{q}}\leq-\dot{\mbm{e}}^*$}, $\vspace{2mm}\begin{bmatrix} -\mbm{J} \\ \mbm{J}
\end{bmatrix}
\dot{\mbm{q}}\leq
\begin{bmatrix}
-\dot{\underline{\mbm{e}}}^* \\ \dot{\overline{\mbm{e}}}^*
\end{bmatrix}$, 
and 
$\begin{bmatrix}
-\mbm{J} \\ \mbm{J}
\end{bmatrix}
\dot{\mbm{q}}\leq
\begin{bmatrix}
-\dot{\mbm{e}}^* \\ \dot{\mbm{e}}^*
\end{bmatrix}$.
If the constraint in~\eqref{eq:ineq_task} is infeasible, a least squares solution for
$\dot{\mbm{q}}$ as in~\eqref{eq:eq_task} can be found by including the slack variable $\mbm{s}$ among the
decision variables and solving the Quadratic Program (QP)
\begin{align}
\begin{split}
&\min_{\dot{\mbm{q}},\mbm{s}} \; \frac{1}{2}\norm{\mbm{s}}^2\\
s.t.~&\mbm{J}\dot{\mbm{q}}\leq\dot{\mbm{e}}^*+\mbm{s}.
\end{split}
\label{eq:QP}
\end{align} 
To enforce a hierarchy of $p = 1, \ldots, P$ priority levels, we stack all task Jacobians of equal priority $p$ in a matrix $\mbm{A}_p$. Also, all corresponding operational space controls $\dot{\mbm{e}}^*$ are stacked in a vector $\mbm{b}_p$ to form one constraint of the form $\mbm{A}_p \leq \mbm{b}_p$ per hierarchy level. The aim is to sequentially satisfy a constraint in the least-square sense while solving for the subsequent constraints of lower priority in the null space of the previous constraint, such that the previous solution is left unchanged. Therefore, the following QP, with the previous slack variable solutions $\mbm{s}_i$ frozen between iterations, needs to be
solved for $p = 1,\ldots,P$
\begin{align}
\begin{split}
&\min_{\dot{\mbm{q}},\mbm{s}_p} \; \frac{1}{2}(\norm{\mbm{s}}^2+\lambda\norm{\dot{\mbm{q}}}^2)\\
s.t.~&\mbm{A}_i\dot{\mbm{q}}\leq\mbm{b}_i+\mbm{s}_i^*,~i=1,\dots,p-1\\
&\mbm{A}_p\dot{\mbm{q}}\leq\mbm{b}_p+\mbm{s}_p.
\label{eq:SoT}
\end{split}
\end{align}
Here, $\lambda \in \mathbb{R}_+$ is used to regularize the solution in order to avoid large velocities due to singularities. The control vector $\dot{\mbm{q}}$ is obtained from the final ($P$-th) solution of (\ref{eq:SoT}).

The main underlying mechanism of the hierarchical framework outlined above is to solve lower ranked tasks as good as possible (in the regularized least-square sense) in the null space of higher ranked task. We exploit this property to incorporate prior knowledge for, $e.\,g.$, obstacle avoidance or desired end-effector alignments. These are posed as tasks on a higher priority level then movement tasks for which control policies are learned as described below. Therefore, no matter what exploration strategy is chosen for learning, by definition the resulting motion can not violate higher prioritized tasks for avoidance and thus a safe behavior is guaranteed. Also, as we will demonstrate in Section~\ref{sec:Exp_eval}, posing additional tasks encapsulating desired behaviors prunes the null space remaining for exploration on lower task levels and thus the policy search space. Therefore, learning in the remaining obstacle free low-dimensional space also facilitates learning rate.

\subsection{Policy search in operational spaces}
\label{sec:RL}
As discussed above, the tasks we consider in this work are described by Jacobian maps from joint space to operational spaces of choice, and corresponding operational space controllers. Here we describe how to learn the corresponding control laws (policies) $\dot{\mbm{e}}^*$ in operational space. Following the work in~\cite{feirstein2016reinforcement}, a natural choice to represent such a policy is a normalized Radial Basis Function (RBF) network

\begin{align}
y(\mbm{x};\mbm{\theta}=\left[\Theta_1,\dots,\Theta_N\right])=\frac{\sum_{i=1}^{N}\Theta_i\rho(\norm{\mbm{x}-\mbm{c}_i})}{\sum_{i=1}^{N}\rho(\norm{\mbm{x}-\mbm{c}_i})},\label{eq:RBFN}
\end{align} 
where $\mbm{x}$ is the state of the system (\textit{e.g} location of the end-effector) and $N$ is the number of basis functions, each of which is centered at individual positions
$\mbm{c}_i$ and weighted by $\Theta_i$. The function 
$\rho(\norm{\mbm{x}-\mbm{c}_i})=\exp(-\frac{\norm{\mbm{x}-\mbm{c}_i}^2}{2\sigma^2})$
 denotes a Gaussian kernel with variance $\sigma$. In this work, $\sigma$ is 
set to $\frac{\text{d}_{\text{max}}}{\sqrt{2N}}$, where $\text{d}_{\text{max}}$ 
is the maximum distance between kernel centers, as this finds a compromise 
between locality and smoothness \cite{benoudjit2002width}. 

A RBF network is essentially a single hidden layer neural
network with $\rho$ as the activation function, $\mbm{x}$ as the input, and $\Theta_i$ as the linear
output weights. It represents a mapping from (potentially high-dimensional) input data to a scalar value
$y:\mbm{x}\in \mathbb{R}^n \mapsto \mathbb{R}$. In this work, the RBF network constitutes a local, time-invariant
policy parameterization that represents operational space policies

\begin{align}
\dot{e}^*_\pi(\mbm{x};\mbm{\theta})=y(\mbm{x};\mbm{\theta}),
\label{eq:dyn_eq}
\end{align}
where $\mbm{x}$ is the input and $\mbm{\theta}$ is the parameter vector. Exploration takes place directly in parameter space 
\begin{align}\label{eq:policy}
\dot{e}^*_\pi(\mbm{x};\mbm{\theta}+\boldsymbol{\epsilon}_t)=y(\mbm{x};\mbm{\theta}+\boldsymbol{\epsilon}_t),
\end{align}
where $\boldsymbol{\epsilon}_t\in\mathcal{R}^{N}$ represent Gaussian noise with variance $\Sigma$. 

The policy parameters $\mbm{\theta}$ are iteratively updated using a time-invariant version of the  PoWER algorithm \cite{kober2009policy}. 
In each iteration, stochastic roll-outs are executed by adding Gaussian noise to the parameter vector. The parameters are then updated as a convex combination of noise and their respective reward, where noise that resulted in higher rewards contributes more to the overall update. An in depth explanation of the algorithm is given in our previous work \cite{lundell2017generalizing}.

Noise is only sampled once per roll-out and added to the most active kernel every time step. To
further control the magnitude of noise, we follow~\cite{lundell2017generalizing} and use a modified
covariance matrix $\Sigma=\gamma\beta\textbf{I}$. Here, $\textbf{I}$ is the identity matrix, $\gamma$ is a
constant controlling the initial magnitude and $\beta$ is given by
\begin{equation}
\label{eq:var_red}
\beta = \frac{1}{\sum_{k=1}^{K}r_k^2}.
\end{equation}
The sum in~\eqref{eq:var_red} is taken over the $K$ best rewards 
$\{r_k\}_{k=1}^K$. 
Parameter $\beta$ essentially
lowers the amount of exploration required once a policy starts converging towards higher rewards.


\section{Experimental Evaluation}
\label{sec:Exp_eval}
We validate our approach by means of two illustrative examples: reaching (Section
\ref{sec:reaching_task}) and grasping (Section \ref{sec:grasping_task}). To this end, we use one of
the $7$ DOF arms of the platform depicted in Fig.~\ref{fig:YuMi}. The reaching experiment was simulated in Gazebo, while the grasping experiment was executed on the actual robot.

The RBF network kernels are evenly distributed over the whole search space. The total number of policy parameters depend on the number of kernels and on the number of control policies to learn, as each policy have their own set of policy parameters and the number of policies grows with the search space dimensionality ($e. \,g.$ one policy for a $1$D search space, two policies for $2$D etc.). Initially all policy parameters are set to zero.

\subsection{Reaching}
\label{sec:reaching_task}
\begin{figure}[t]
	\vspace{3mm}
	\centering
	\begin{subfigure}{.25\textwidth}
		\centering
		\vspace{-1.5em}
		\includegraphics[width=1\linewidth]{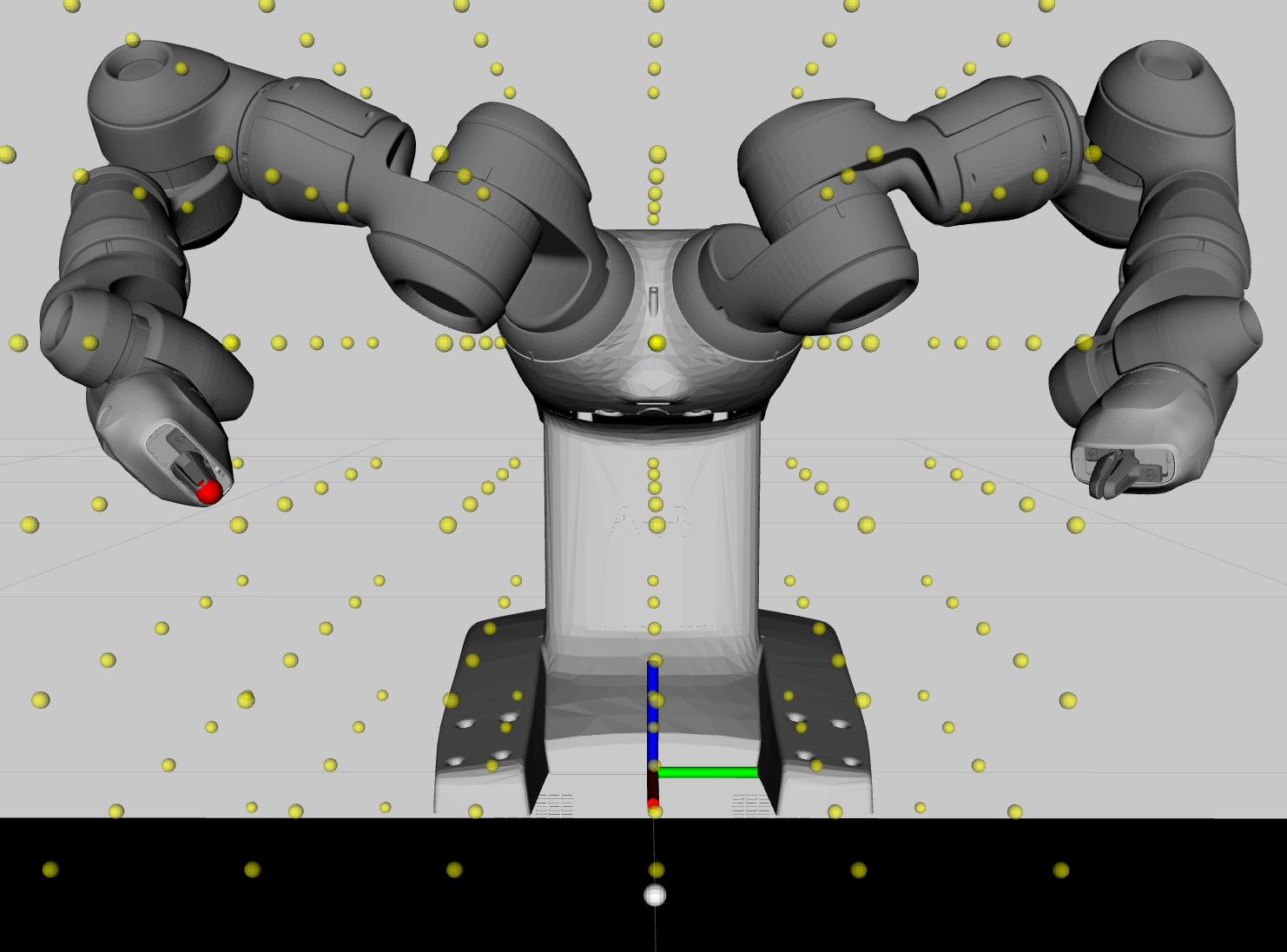}
		\caption{Only the reaching task is specified.}
		\label{fig:YuMi_sim_grid_no_task}	
	\end{subfigure}%
	\begin{subfigure}{.23\textwidth}
		\centering
		\includegraphics[width=.95\linewidth]{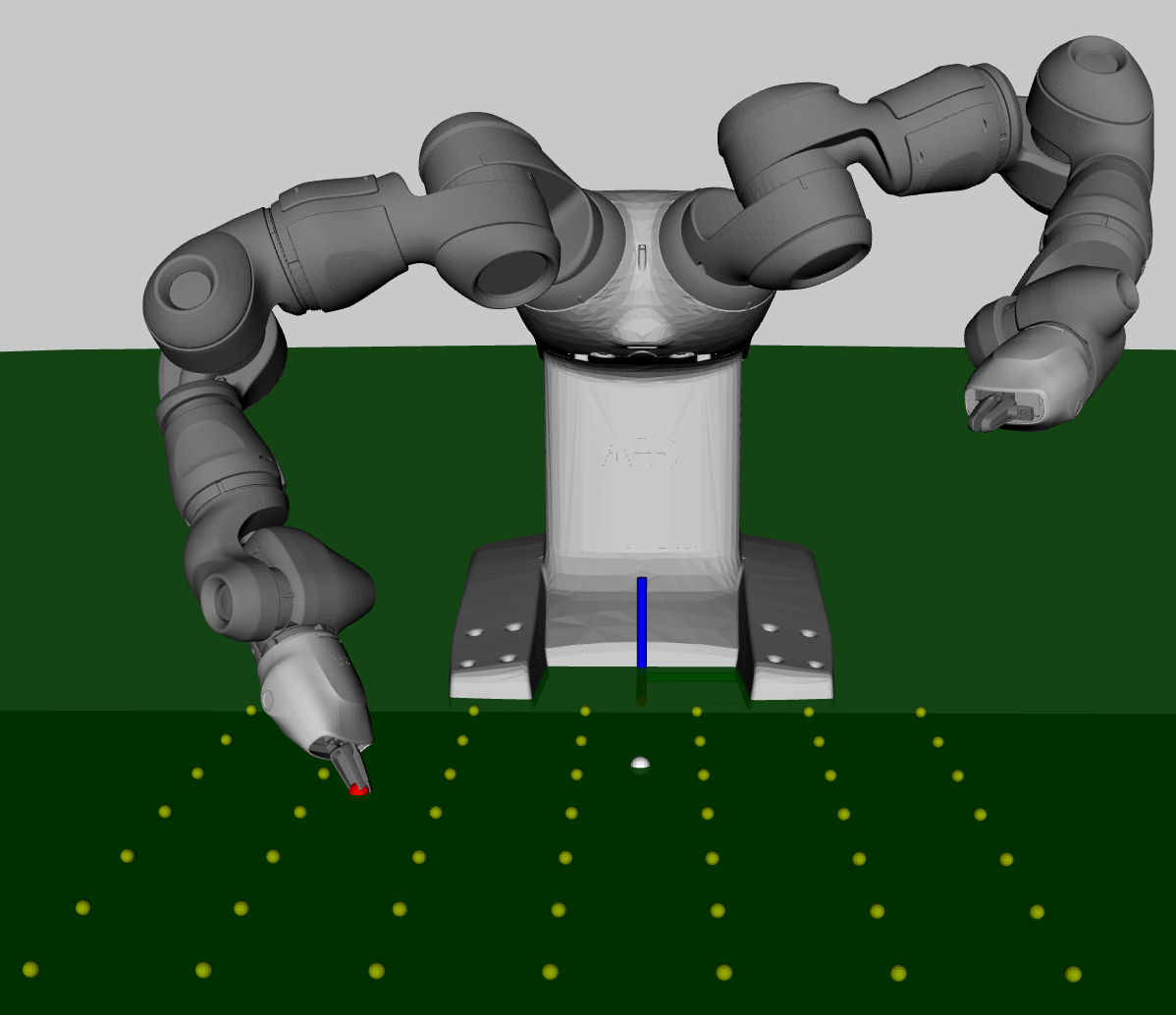}
		\caption{In addition to the reaching task a projection task forcing the 
		red sphere onto the green plane is added.}
		\label{fig:YuMi_sim_grid_proj_task}	
	\end{subfigure}
	\caption{Reaching experiment: The yellow spheres represent the Gaussian 
	kernels
		in~\eqref{eq:RBFN}, the red sphere the end-effector point and the white 
		sphere the target
		position. The black surface indicates an obstacle. (Best viewed in 
		color).}
	\label{fig:YuMi_sim}
		  \vspace{-5mm}
\end{figure}

In this experiment, the main goal is to analyze the impact optional higher ranked tasks have on safety when learning movement policies on lower priority levels. As a side objective, we also want to study the effect on learning rate when adding additional higher ranked tasks that are not required for ensuring safety. To achieve this we chose a reaching experiment where the objective for the robot, illustrated in Fig.~\ref{fig:YuMi_sim}, is to colocate the red end-effector point \mbm{p} with the white target point \mbm{k} defined in the environment. Moreover, to actually validate the impact higher ranked tasks have on the learned policy, the experiment was split in three sub-experiments with varying prior knowledge.

In the first sub-experiment (Exp. 1), shown in Fig.~\ref{fig:YuMi_sim_grid_no_task}, no additional tasks were encoded besides the reaching
task. Consequently, the STESS defaulted to
$3$D Cartesian space including the black obstacle. The second sub-experiment (Exp. 2) extends Exp. 1
by encoding a task that avoids obstacles on a higher priority level but maintains the same search
space. In addition to the constraints in Exp. 2, the final sub-experiment (Exp. 3), shown in
Fig.~\ref{fig:YuMi_sim_grid_proj_task}, encodes a task that forces end-effector point $\mbm{p}$ to
also lie on the green plane, effectively reducing STESS from $3$D to $2$D.

Here, the operational space of the reaching task, posed at the lowest hierarchy level, corresponds
to $3$D Cartesian space. Thus, the corresponding task Jacobian in~\eqref{eq:ineq_task} is simply the
manipulator Jacobian with respect to $\mbm{p}$. For each Cartesian dimension, an
operational space policy is learned where the state $\mbm{x}$ in~\eqref{eq:policy} corresponds to the end-effector position.

Each experiment ran for $10$ trials with a maximum of $300$ roll-outs (including $15$ initial roll-outs before policy update) per trial. The importance sampler used the five best roll-outs for each policy update. If a collision occurred during training, the corresponding trial was deemed unsuccessful. The immediate reward function at time $t$ is
\begin{equation}
\label{eq:rew_reaching}
r(\dot{\mbm{q}}_t,\mbm{p}_t) = 
\begin{cases}
\exp\left(-\alpha_1\norm{\dot{\mbm{q}}_t}_1-\alpha_2d^2\right),&\text{if}\ t=\text{T}\\
\exp\left(-\alpha_1\norm{\dot{\mbm{q}}_t}_1\right), & \text{otherwise}
\end{cases}
\end{equation}
where $\dot{\mbm{q}}$ are joint velocities and $\mbm{p}_t$ is the 3D Cartesian position of the gripper. 
The term $d=\norm{\mbm{p}_{\text{T}}-\mbm{k}}_2$ is the distance between the 
gripper \mbm{p} and the 
target point \mbm{k} at the end of the roll-out. The parameters $\alpha_j$ are individual weighting 
factors chosen as $\alpha_1=0.001$ and $\alpha_2=10$. 

In Exp. 1 and 2 we chose $249$ kernels and, with a $3$D search space, the total number of policy
parameters were $747$. In Exp. 3, on the other hand, the search was carried out in $2$D and
therefore we chose only $49$ kernels, resulting in $98$ policy parameters. For Exp. 1 and 2, the
initial exploration parameter $\gamma=4$, while for Exp. 3 $\gamma=0.001$. The parameter $\beta$
controlling the noise level was calculated using the reward from the $10$ best roll-outs.

\begin{figure}
	\centering
	\includegraphics[width=1\linewidth]{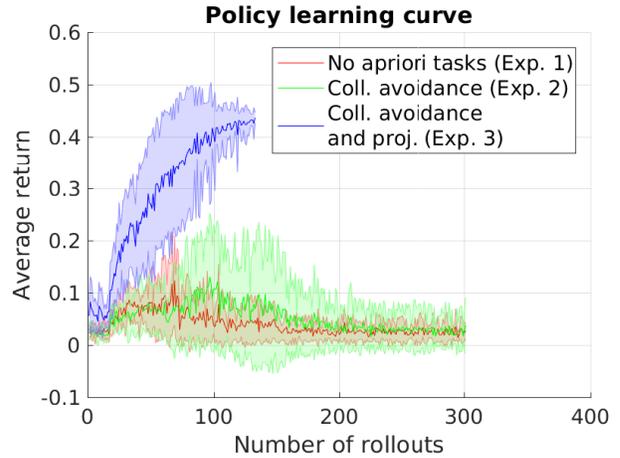}
	\caption{The average return over the number of roll-outs for each experiment.}
	\label{fig:simulater_rewards}	
	\vspace{-3mm}
\end{figure}

Figure \ref{fig:simulater_rewards} shows the convergence rates of Exp. 1, 2, and 3. These, together
with the data presented in Table \ref{tab:collisions}, indicate that obstacle avoidance eliminates
the possibility for collisions during exploration, but does not improve the actual learning
rate. However, encoding additional tasks that reduced the search space to $2$D allowed the policy to
converge in $100$\% of the trials after an average of $84.4$ roll-outs. 

The main objective in this experiment was to study the implication higher ranked tasks have on learning safety and, as the results indicate, our method is indeed capable of forming STESS that are collision free. However, based on the results from Exp. 1 and 2, it seems impossible to learn RBF network policies that successfully fulfill the task in high dimensional search space, probably originating from poor scaling of RBF networks to high dimensional state spaces \cite{friedman2001elements}. In such cases special handcrafted policy representations are preferred \cite{peters2007reinforcement,ijspeert2002movement,kalakrishnan2011learning,feirstein2016reinforcement}. Nevertheless, as indicated in Exp. 3, adding additional domain knowledge projecting STESS to a lower dimensionality allowed general RBF network policies to converge in all trials. In terms of achieving higher learning rate it is possible to use other policy search algorithms as our method do not default to any particular one. Then again, based on the results in Exp. 3, the most significant learning boost is achieved by further limiting the STESS with additional constraints, something which is further demonstrated in the following grasping experiment.

\begin{table}[]
	\centering
  \def\arraystretch{1.1}
	\begin{tabular}{l|l|l|l}
		& Exp. 1 & Exp. 2 & Exp. 3 \\ \hline
		\begin{tabular}[c]{@{}l@{}}Collision during\\ training\end{tabular} & 
		\mbm{4/10~(40\%)}& \mbm{0/10~(0\%)} & 0/10 
		(0\%) \\ \hline
		Converged trials          & 0/10 (0\%) & 0/10 (0\%) & 
		\mbm{10/10~(100\%)} 
	\end{tabular}
	\caption{Collision frequency and learning performance.}
	\label{tab:collisions}
	  \vspace{-3mm}
\end{table}

\subsection{Grasping}
\label{sec:grasping_task}

\begin{figure}
	\centering 
	\includegraphics[width=0.35\textwidth]{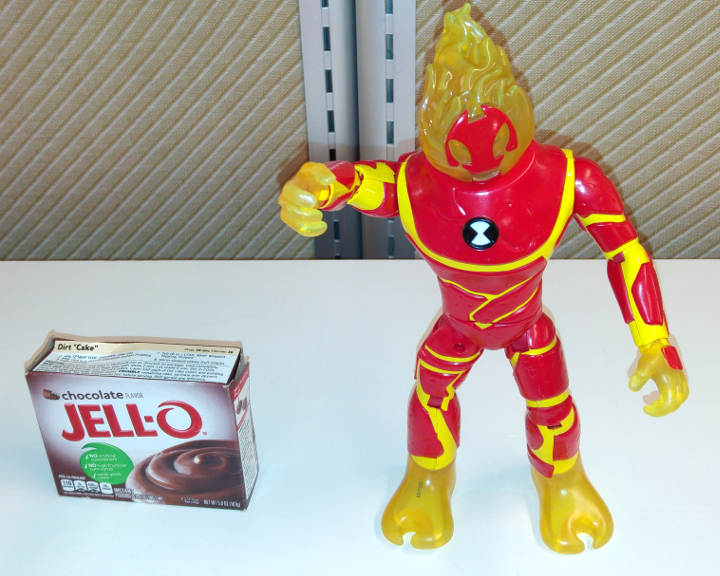}
	\caption{Shows the two test objects used in the grasping experiment. The robot could only grasp the toy at specific heights from the smaller edges, whereas the box was graspable at all heights from shorter edges.}
	\label{fig:box_and_toy}	
	\vspace{-6mm}
\end{figure}

In this experiment, the goal is to demonstrate our method working on a real robot in the sense that it learns policies for grasping nontrivial objects after few roll-outs. To achieve this, the experiment is to grasp the toy and box displayed in Fig.~\ref{fig:box_and_toy} with the robot shown in Fig.~\ref{fig:YuMi}. The
results presented in the previous section indicate that prior data in form of additional tasks
allows for both faster and safer policy search. Therefore, we attempt to maximize the amount of
prior knowledge in order to reduce the search space. To this end, we use tasks forming a
so-called grasping envelope as defined in our previous work~\cite{stoyanov2016grasp}. These grasping
envelopes are intended to capture grasping strategies observed in
humans~\cite{balasubramanian2012physical}. In~\cite{stoyanov2016grasp} we used them to successfully
produce robust grasps for a wide selection of objects including bottles, boxes, and plush toys.

\begin{figure}
	\centering
		\begin{subfigure}{0.30\textwidth}
			\centering \includegraphics[scale=0.4]{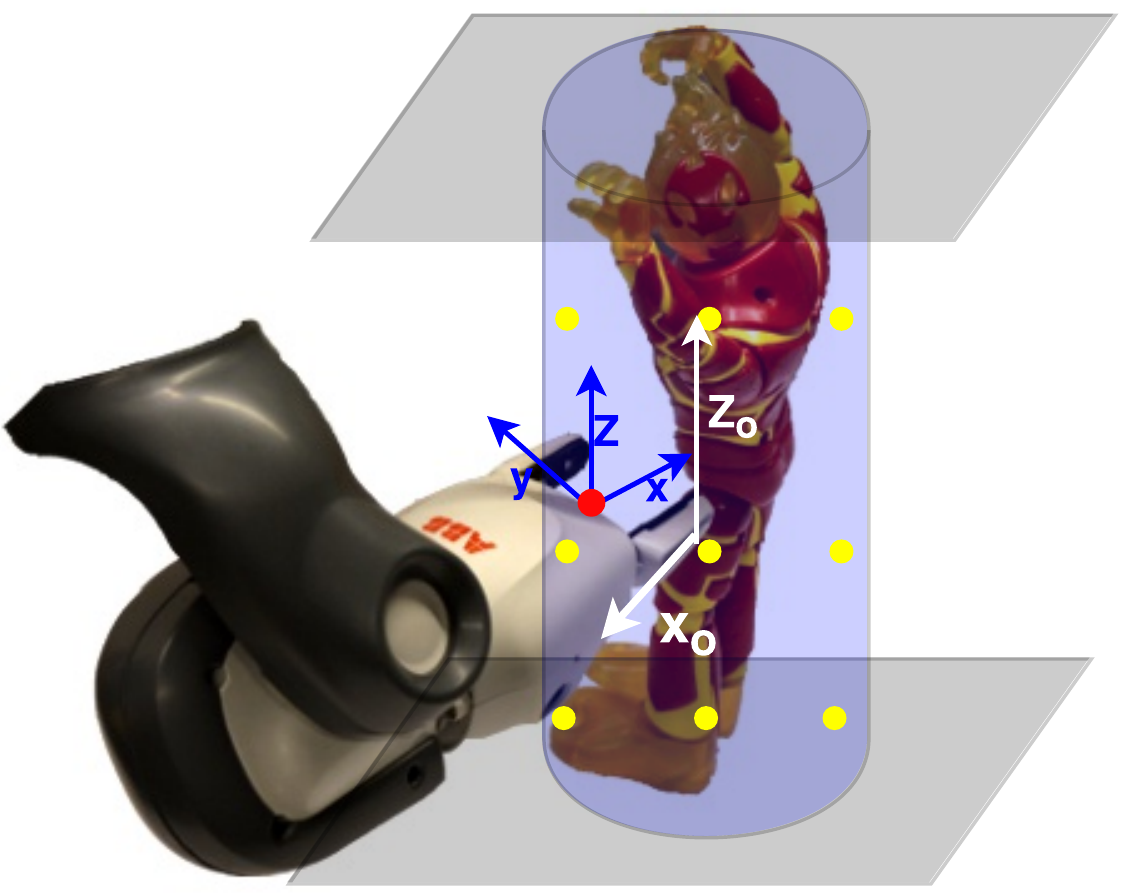}
			\caption{}
			\label{fig:manifold_outer3d_shell}	
		\end{subfigure}
	\begin{subfigure}{0.5\columnwidth}
		\centering
		\includegraphics[scale=0.215]{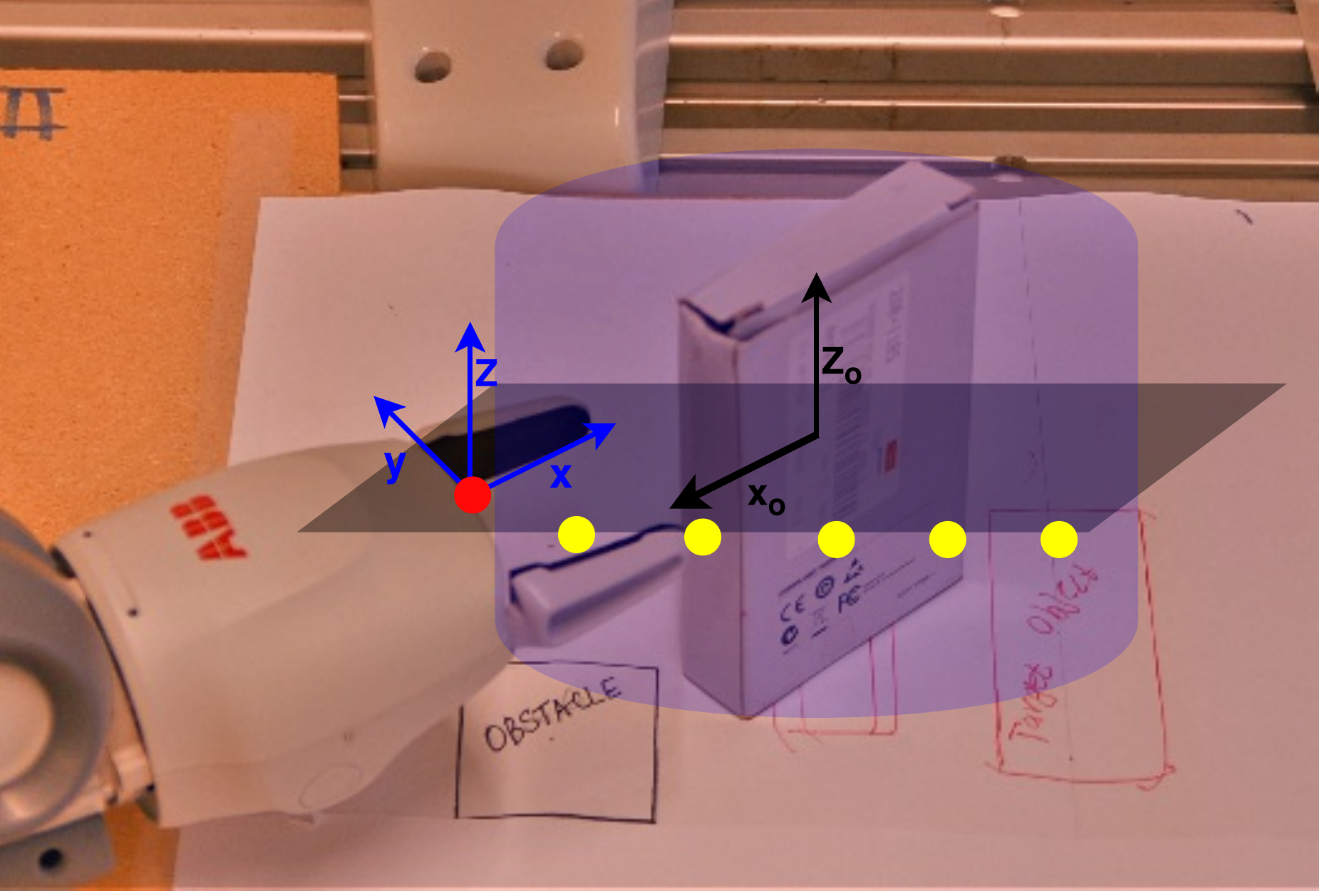}
		\caption{}
		\label{fig:manifold_outer_shell}	
	\end{subfigure}%
		\begin{subfigure}{0.27\textwidth}
			\centering \includegraphics[scale=0.3]{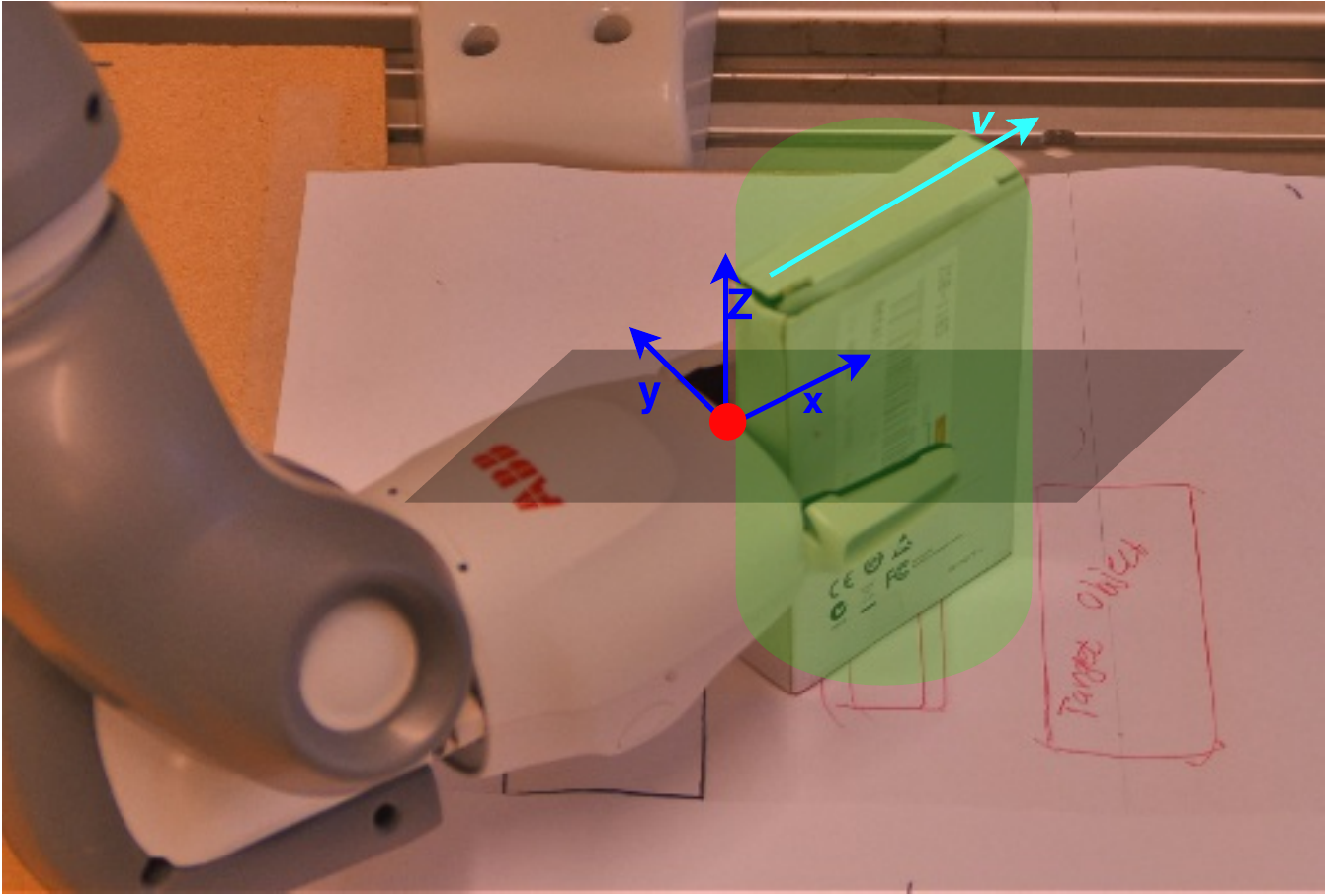}
			\caption{}
			\label{fig:manifold_inner_shell}	
		\end{subfigure}
	\caption{Grasping experiment: (a) The grasp manifold consists of a larger blue cylinder and the black grasping planes. The red end-effector point $\mbm{p}$ is forced to lie on the blue
          cylinder and between the black planes in (a) and on the black plane in (b). In (c) the red end-effector point is constrained to lie on the smaller green cylinder. The $x$ and
          $z$ axis of the gripper are aligned with the corresponding $x_o$ and $z_o$ axis of the
          cylinder. In (a) and (b), the yellow points indicate the RBF kernel distribution
          in~\eqref{eq:RBFN}. Furthermore, one principal component of
          the target object is indicated by the cyan vector $\mbm{v}$ in (c).}
	\label{fig:simplified_manifold}
	\vspace{-5mm}
\end{figure}

In this experiment orientation constraints of the gripper are consistent with the grasping envelope in~\cite{stoyanov2016grasp}, that is: the gripper's vertical axis $z$ (see Fig.~\ref{fig:simplified_manifold}) is enforced to align with the cylinder axis $z_0$, while
 the gripper's approach axis $x$  is to point towards the cylinder axis $x_0$. 
 Moreover, during the roll-out the end-effector point $\mbm{p}$ is constrained 
 to lie on a larger blue cylinder as shown in 
 Fig.~\ref{fig:manifold_outer_shell}, while after the roll-out the same 
 $\mbm{p}$ is constrained to the smaller green cylinder displayed in 
 Fig.~\ref{fig:manifold_inner_shell}, thus allowing the gripper to enclose the 
 object. Finally, when grasping the toy in 
 Fig.~\ref{fig:manifold_outer3d_shell} the gripper is constrained to lie 
 between 
 the upper and lower black planes, while for grasping the box in 
 Fig.~\ref{fig:manifold_outer_shell} it is constrained to lie on a single black 
 plane. Together these constraints leave redundancy for the gripper to move 
 both vertically and horizontally around the manifold when grasping the toy, 
 \textit{i.e.} a $2$D search space, and only horizontally when grasping the 
 box, effectively reducing the search space from $2$D to $1$D.

The movement tasks forcing the gripper to converge onto the grasp manifold were predefined using simple controllers of the form $\dot{\mbm{e}}^*=-\lambda \mbm{e}$. On the lowest priority level (and thus in the null space of all higher ranked task), we learned policies modulating the approach motion. In the case of grasping the toy, the aforementioned predefined reaching and alignment tasks leave the end-effector free to move in an operational space which is tangent to the cylinder shown in Fig.~\ref{fig:manifold_outer3d_shell}. Therefore, to learn a policy modulating the movement in this redundant space, we define a task whose Jacobian maps to the cylinder's tangent space. In the case of grasping the box, the redundant space is further constrained to the pre-defined grasping plane illustrated in Fig.~\ref{fig:manifold_outer_shell}. Thus, in this case we form an operational space mapping via a corresponding Jacobian which maps to the tangent space of the cylinder lying in the plane. When learning the modulation policy, exploration happens then only in this reduced redundant space. Based on this, the state $\mbm{x}$ in~\eqref{eq:policy} when grasping the toy is the $(x,y,z)$ position of the end-effector while for the box reduces to $(x,y)$ position.

Due to the limited opening size of the gripper, it could only grasp the toy and the box in Fig.~\ref{fig:box_and_toy} close to the shorter edges. Additionally the odd shape of the toy made it graspable only at specific heights, further complicating the situation. Hence, a suitable reward function for both objects needs to guide the learning towards either of the short sides. We devised such a reward function by reusing~\eqref{eq:rew_reaching}. Here however, the parameter $d$ represents the shortest distance
between end-effector point $\mbm{p}$ and a vector $\mbm{v}$ passing parallel to one of the principal components
of the object (see Fig.~\ref{fig:manifold_inner_shell}). The reward function for grasping is given by
\begin{align}
r(d,I) = \exp\left(-\eta_1d-\eta_2I\right)
\label{eq:rew_grasping}
\end{align}
where $I$ is a binary variable indicating a successful (0) or unsuccessful (1) grasp. The term $d$ is calculated as $d=\norm{\mbm{p}_T-\mbm{v}}$ where $\mbm{p}_T$ is the 3D Cartesian position of the gripper at the end of the roll-out. The parameters $\eta_1$ and $\eta_2$ were, respectively, set to $700$ and $1$ when grasping the box, and $7$ and $1.5$ when grasping the toy. Grasp failure is detected if the gripper opening joint value crosses a predefined threshold indicating an empty grasp. 

In this work we predefined a principal component as the preferred grasp direction $\mbm{v}$, but it could also be determined using principal component analysis on a discrete representation of the target object's geometry. This, however, we considered not to be in the scope of the presented work and the literature offers many techniques to accomplish this task \cite{liu2009robust}. The reward function effectively guides the grasp towards one principal component of the object which, based on human grasping strategies, have been demonstrated to produce robust grasps \cite{balasubramanian2012physical}. Although the reward function is not the core contribution in this work, it is potentially useful for other learning tasks.

For the grasping experiment the number of trials, maximum roll-outs, and number of roll-outs used by the importance sampler was the same as in the previous experiments. The number of initial roll-outs before policy iteration was set to $8$. By trial and error, we found that $24$ kernels were enough to learn a good policy for grasping the toy and $11$ for grasping the box. As the search space was $2$D for the toy and $1$D for the box the number of policy parameters were, respectively, $48$ and $11$. The exploration parameter $\gamma$ when grasping the toy was $0.0006$ for kernels modulating vertical ($z$) movements and $0.00009$ for modulating horizontal ($x$ and $y$) movements. The same exploration parameters when only learning horizontal movements for grasping the box was set to $0.001$. In both cases the seven best rewards were used to calculate $\beta$.

\begin{figure}
	\centering
	\includegraphics[width=0.95\linewidth]{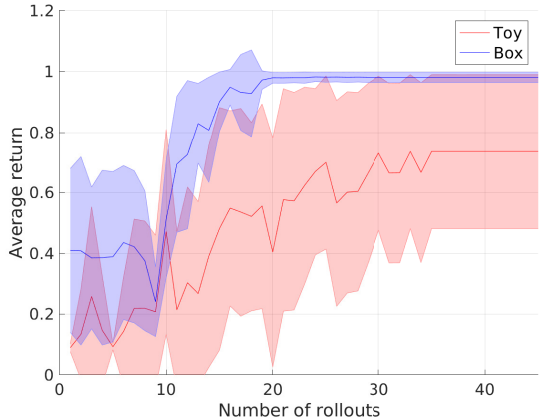}
	\caption{The average return over the number of roll-outs for grasping the box and the toy. The reason the rewards for grasping the toy and box differs so much is because of different weighing parameters $\eta_j$ in (\ref{eq:rew_grasping}).}
	\label{fig:grasping_rewards}	
	\vspace{-5mm}
\end{figure}

\begin{figure*}%
	\centering
	\begin{subfigure}[b]{.4\columnwidth}
		\includegraphics[width=1.06\columnwidth]{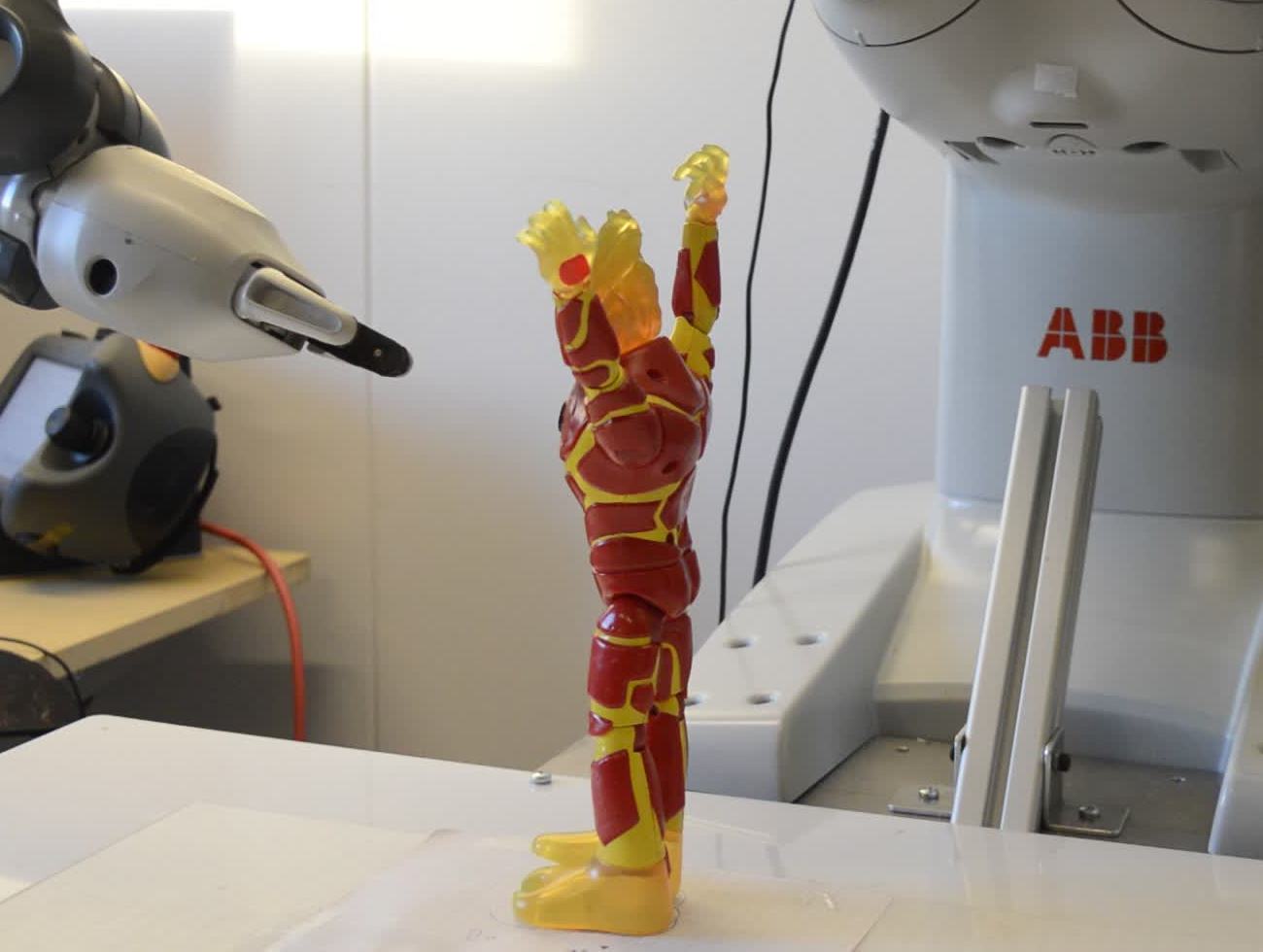}%
	\end{subfigure}
	\begin{subfigure}[b]{.4\columnwidth}
		\includegraphics[width=1.06\columnwidth]{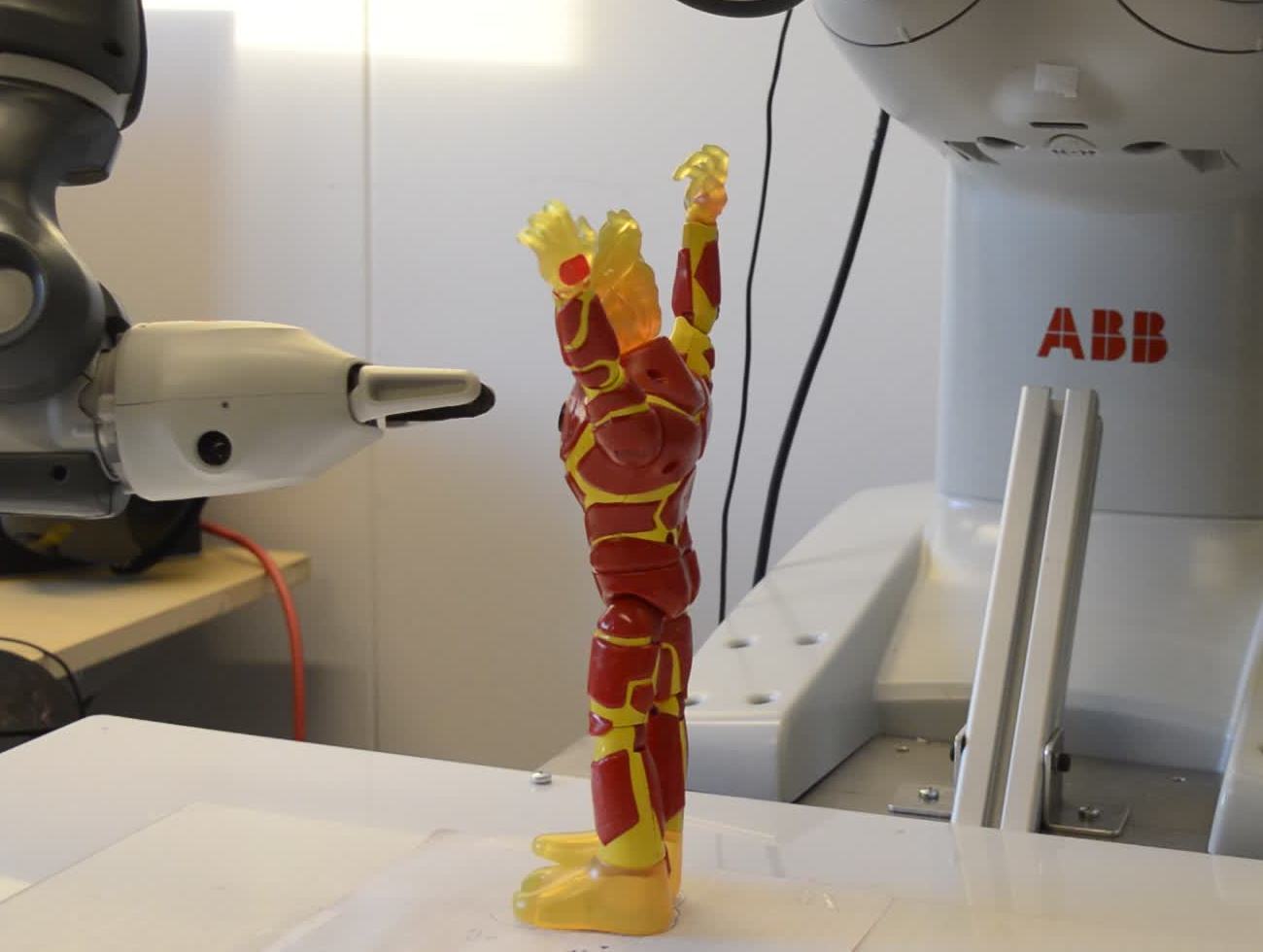}%
	\end{subfigure}
	\begin{subfigure}[b]{.4\columnwidth}
		\includegraphics[width=1.06\columnwidth]{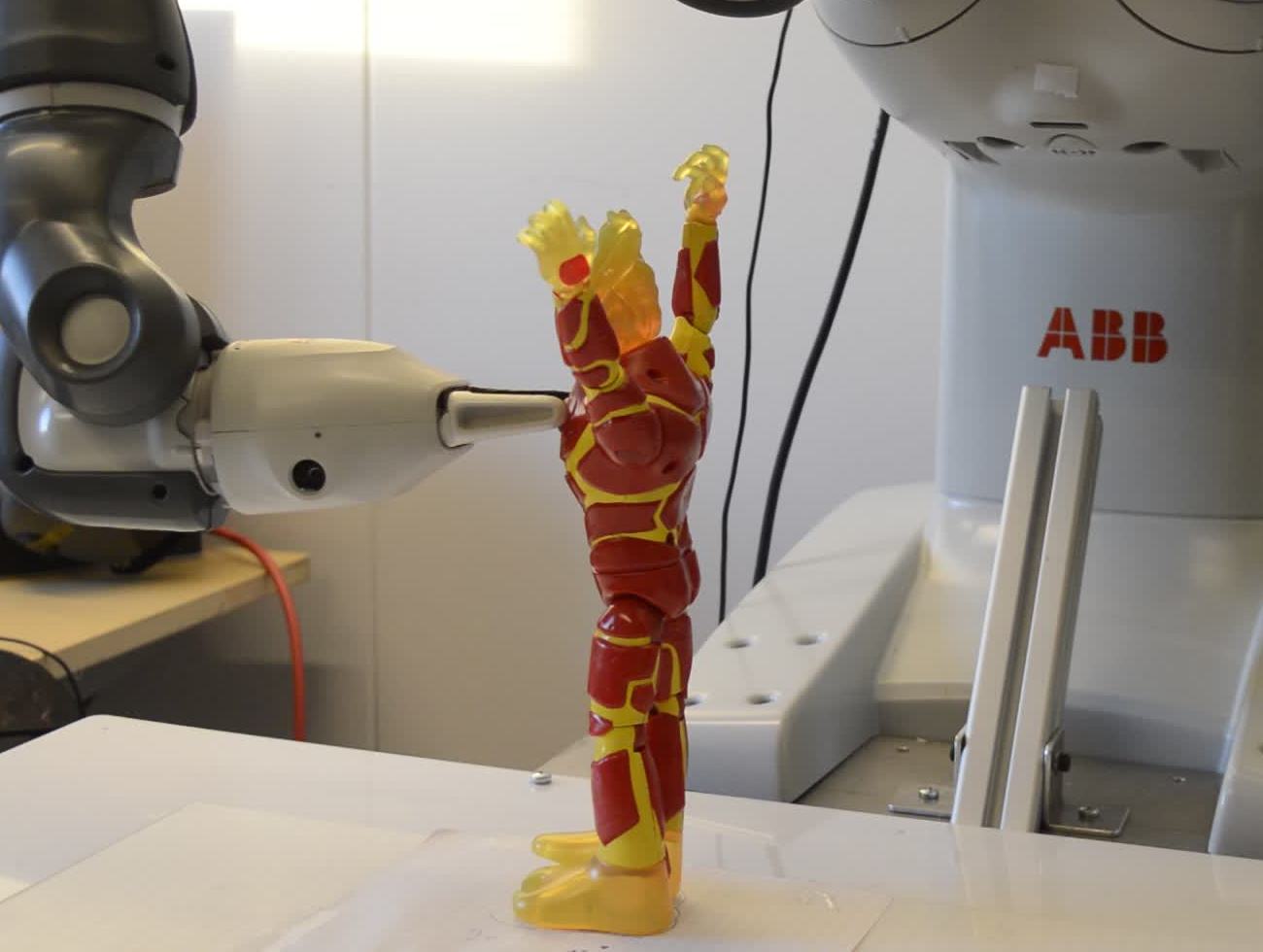}%
	\end{subfigure}%
	\begin{subfigure}[b]{.4\columnwidth}
		\includegraphics[width=1.06\columnwidth]{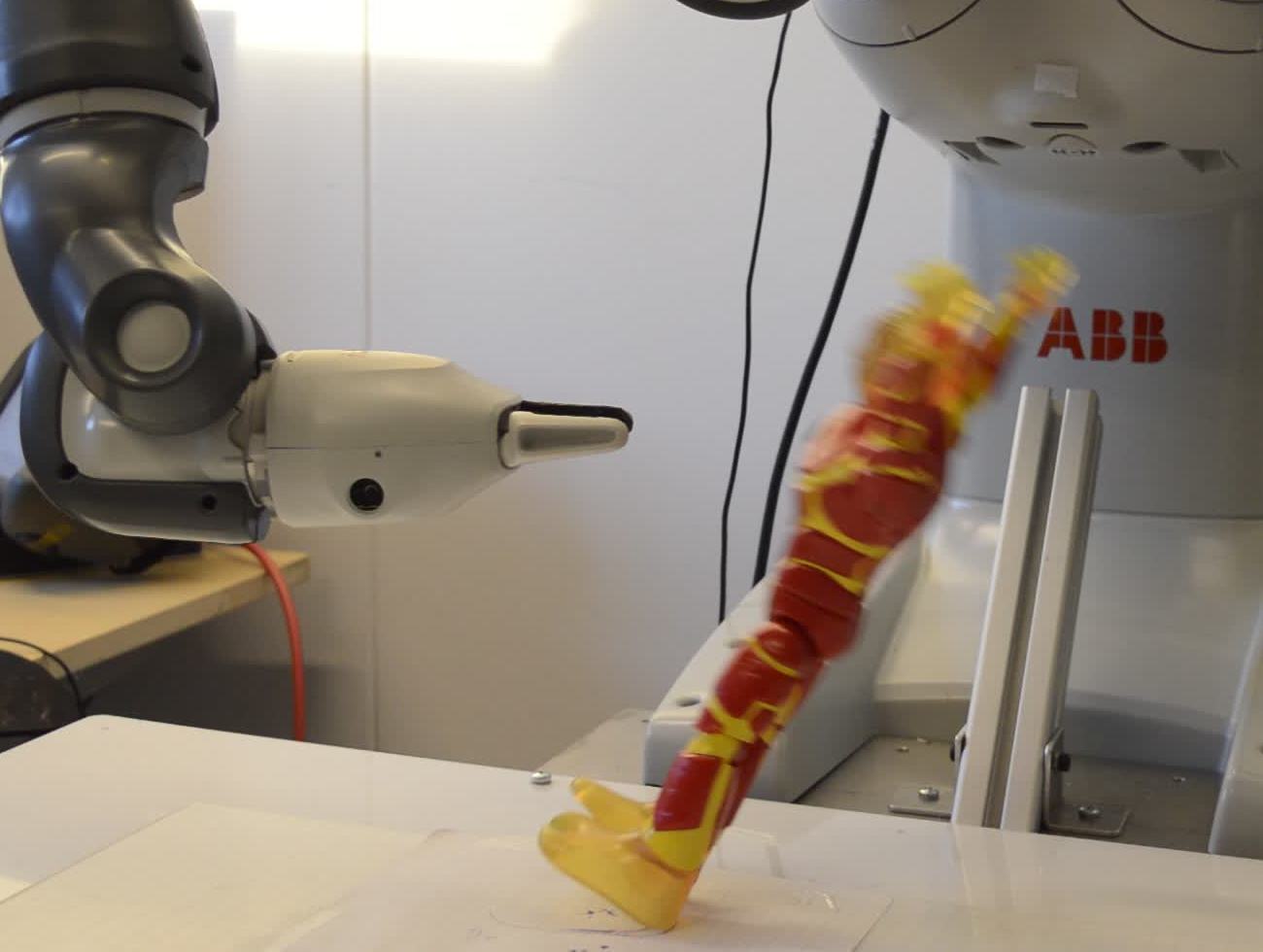}%
	\end{subfigure}%
	\\	
	\begin{subfigure}[b]{.4\columnwidth}
		\includegraphics[width=1.06\columnwidth]{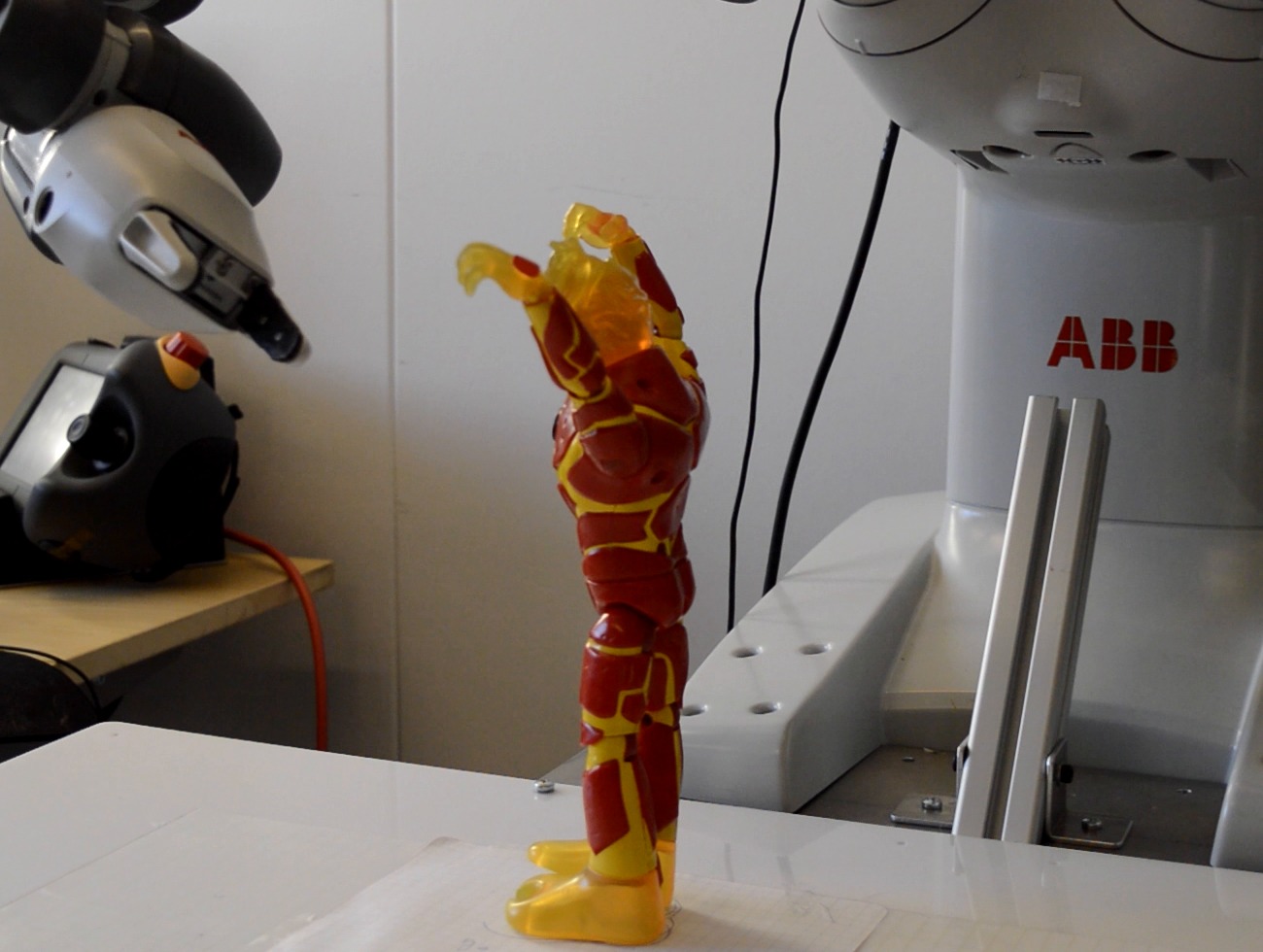}%
	\end{subfigure}
	\begin{subfigure}[b]{.4\columnwidth}
		\includegraphics[width=1.06\columnwidth]{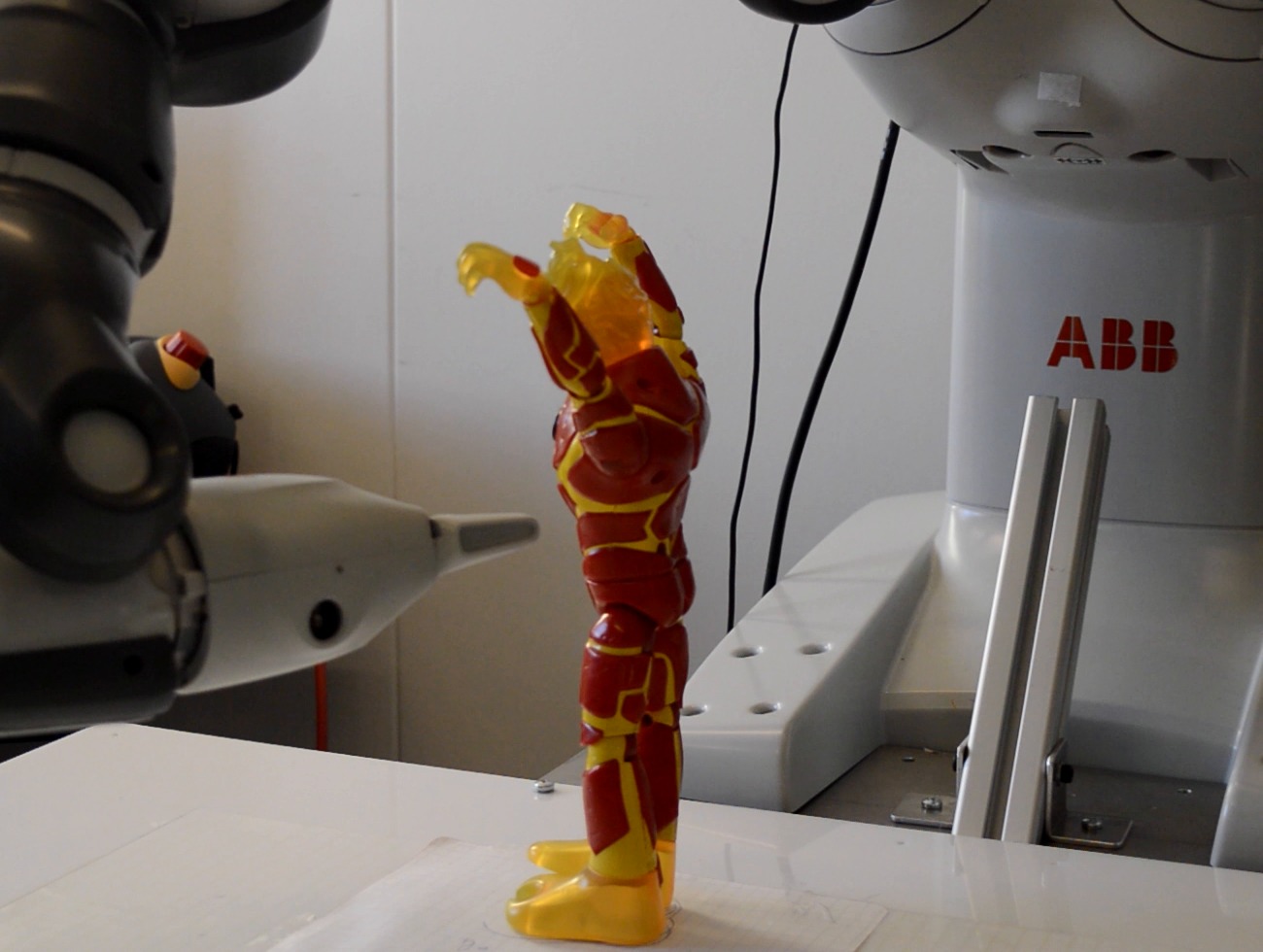}%
	\end{subfigure}
	\begin{subfigure}[b]{.4\columnwidth}
		\includegraphics[width=1.06\columnwidth]{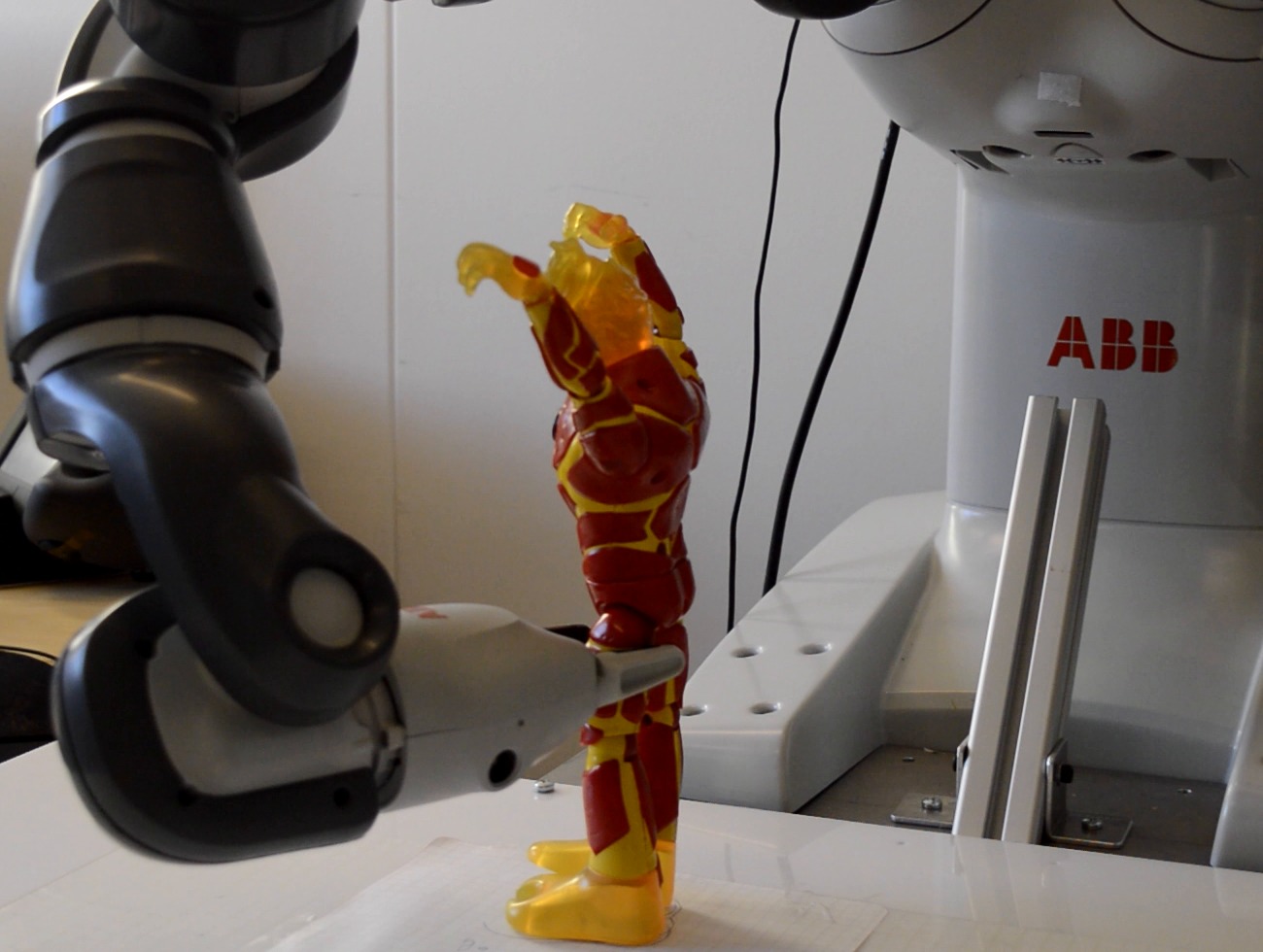}%
	\end{subfigure}%
	\begin{subfigure}[b]{.4\columnwidth}
		\includegraphics[width=1.06\columnwidth]{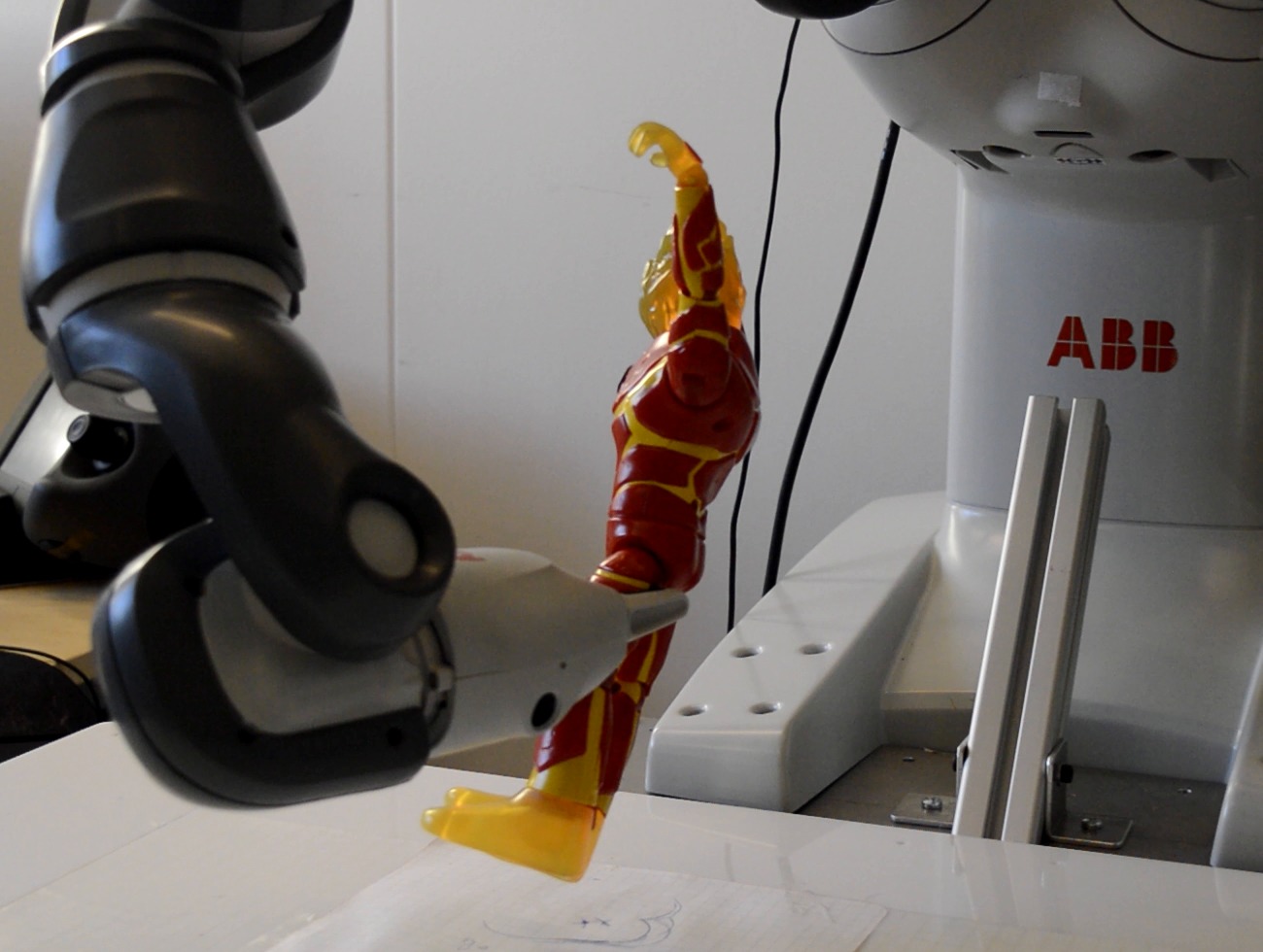}%
	\end{subfigure}%
	\caption{Sequence of images from execution of the initial unlearned policy (top figure) and the final learned policy (bottom figure) for graping the toy. It is clearly visible that grasping the toy requires precise vertical and horizontal placement of the manipulator.}
	\label{fig:grasp_motion}
	\vspace{-5mm}
\end{figure*}

The convergence rate of the policy search is illustrated in Fig.~\ref{fig:grasping_rewards}. In both cases the
policy converged in all $10$ trials after $16.3$ roll-outs on average when grasping the box and $39$ when grasping the toy (including $8$ initial roll-outs before policy optimization), after this the robot was able to consecutively grasp the objects. Fig.~\ref{fig:grasp_motion} shows an example executing the initial unlearned policy as well as the final learned policy. 

The results clearly indicate that forming STESS with our method enables a real robot to learn both safely and efficiently to grasp even very complex objects. As already mentioned in the reaching experiment adding additional tasks boost the learning significantly, a claim further strengthened by the $49$\% reduction in number of roll-outs before convergence when learning to grasp the box opposed to the toy. Although part of the fast learning rates presented in Fig.~\ref{fig:grasping_rewards} originates from well tuned meta-parameters such as exploration rate $\gamma$ and decay $\beta$, they are not crucial for the method to work, as poorly tuned parameters will not prohibit learning but instead just slow it down. In conclusion, the most important factor for learning is prior knowledge added to the system in the form of higher ranked tasks, and by adding tasks intended to mimic human grasping strategies not only reduced the search space to a collision free low-dimensional representation, but also led to successful grasping. 

\section{Related Work}
\label{sec:related_works}
The core ingredients of our work are safe policy search, operational space learning and the
normalized RBF network policy. Similar ideas for safe exploration as presented 
here was recently studied in \cite{pham2018optlayer} where they force output 
actions from a neural network to safe spaces by taking closest actions that 
satisfy specific constraints. In that work, however, they only constrained a 
single QP at each iteration while we impose hierarchical constraints allowing 
greater flexibility. 
Other 
methods trying to ensure safe exploration enforce conservative
policy updates between iterations
\cite{peters2008reinforcement,schulman2015trust,ammar2015safe,peters2010relative,kober2009policy}. These
methods are typically combined with low-dimensional policy representations that are initialized on
prior data, $e.\,g.$, human demonstrations or trajectories optimized using an initial
model~\cite{kalakrishnan2011learning,feirstein2016reinforcement,ijspeert2002movement}. Limiting
policy updates between iterations lowers the probability of straying into unexplored state space,
but does not provide any guarantees~\cite{garcia2012safe}. The associated risk is also evidenced by
the results reported in Section~\ref{sec:Exp_eval}: if the search space includes obstacles,
collisions can and, in general, will occur between the robot and obstacles. Another approach to safe
behavior is to directly discourage entering part of the state space by inferring
penalties~\cite{deisenroth2011learning}. These penalties are incorporated as a term in the reward
function that punishes the robot if it comes in proximity to obstacles. These method bears
resemblance to classical potential field methods to obstacle avoidance~\cite{Khat86}. Similar to
potential fields, while significantly reducing the collision rate, state space penalization does not
provide guarantees while introducing additional tuning parameters. In comparison, our method can
completely remove the collision risk during exploration without additional tuning parameters by
posing appropriate avoidance tasks.

Regarding the policy representation used in this paper, previous work in value-based RL uses RBF
networks to approximate the value function~\cite{kretchmar1997comparison}, while in policy search it
is a vital part of the popular Dynamic Movement Primitives~\cite{ijspeert2002movement}. Only
recently was it used as a standalone policy representation optimized to learn a limit-cycle walking
gait in simulation \cite{feirstein2016reinforcement}. This, in addition to our work, indicates that RBF networks are applicable policy representations for learning a variety of tasks. 

In terms of operational space learning, we can also consider prior methods that learn the maps
from joint space to operational space ($i.\,e.$, Jacobians) together with the corresponding control
laws~\cite{peters2008learning}. In that work, the controller is represented as a locally linear 
policy which
is optimized with reward weighted regression. If no prior information regarding a skill is known,
learning the Jacobians and potentially the dynamics in operational space can be valuable. It would
be interesting to extend our framework to account for these possibilities.


\section{Conclusion and Future Work}
\label{sec:discussion}
\balance
We presented a method for safe and sample-efficient learning of policies in arbitrary operational
spaces. The key concept is a method to implicitly form \textit{safe-to-explore state spaces} for policy search by decomposing a skill into elemental sub-tasks focusing exploratory actions to a collision free subspace of the original search space. By decomposing a skill of interest into elemental sub-tasks,
our method allows to encode prior knowledge in form of task constraints. Movement policies are
learned in the null space of higher ranked tasks which enforce, \textit{e.g.}, obstacle avoidance or
desired end-effector alignments. Thus, by construction, our approach ensures safe exploration. Also,
the size of the search space available for learning task policies can be controlled by posing higher
ranked tasks which implicitly prune the redundant space remaining for exploration. We provide an
experimental evaluation by means of a simulated reaching task demonstrating that our method allows
safe, collision-free movement policy learning while simultaneously leading to an increased learning
rate. Furthermore, we evaluated our approach via grasping tasks executed on a real robot, achieving
fast policy convergence even for complex objects.

To speed up learning, we predefine some parameters (kernel placement and width) of the chosen policy
representation. This limits the applicability of the method, but also opens up interesting future
research avenues. One option is to incorporate visual feedback of the scene~\cite{stoyanov2016grasp}
in order to decide on the number of kernels and their individual placement. Another option is to
treat kernel centers and widths as additional policy parameters and to learn them using model-free
policy search. Also, while our framework is not tied to a specific policy representation, the
policies learned in this work are local. Therefore, they will not generalize well to new unseen
situations, such as different object rotations. To allow generalization, one 
option is to learn
global policy representations such as (deep) neural networks. Also, it would be 
interesting to train
several local policies in varying settings and to combine them into a global 
model allowing
inter- and extrapolation~\cite{lundell2017generalizing}.


\bibliographystyle{IEEEtran}
\bibliography{references}

\end{document}